\documentclass[sn-mathphys]{sn-jnl}% Math and Physical Sciences Reference Style
\jyear{2022}%
\usepackage{array}
\usepackage{mdwtab}
\usepackage{eqparbox}
\usepackage{url}
\usepackage{booktabs}
\usepackage{subcaption}
\usepackage{float}
\usepackage{lipsum}
\usepackage{soul}
\usepackage{lineno,hyperref}
\theoremstyle{thmstyleone}%
\theoremstyle{thmstyletwo}%

\theoremstyle{thmstylethree}%

\begin{document}

\title[Deepfake Detection of Occluded Images]{Deepfake Detection of Occluded Images Using a Patch-based Approach}


\author[1]{\fnm{Mahsa} \sur{Soleimani}}\email{mah.soleimani@mail.sbu.ac.ir}
\author[1]{\fnm{Ali} \sur{Nazari}}\email{al\_nazari@sbu.ac.ir}
\author*[1]{\fnm{Mohsen} \sur{Ebrahimi Moghaddam}}\email{m\_moghadam@sbu.ac.ir}

\affil*[1]{\orgdiv{Faculty of Computer Science and Engineering, Shahid Beheshti University}, \orgaddress{\street{Evin}, \city{Tehran}, \postcode{1983969411},  \country{Iran}}}
%\affil[2]{\orgdiv{Shahid Beheshti University}, \orgaddress{\street{Evin}, \city{Tehran}, \postcode{1983969411},  \country{Iran}}}

\abstract{
DeepFake involves the use of deep learning and artificial intelligence techniques to produce or change video and image contents typically generated by GANs. Moreover, it can be misused and leads to fictitious news, ethical and financial crimes, and also affects the performance of facial recognition systems.
Thus, detection of real or fake images is significant specially to authenticate originality of people's images or videos. One of the most important challenges in this topic is obstruction that decreases the system precision.
In this study, we present a deep learning approach using the entire face and face patches to distinguish real/fake images in the presence of obstruction with a three-path decision: first entire-face reasoning, second a decision based on the concatenation of feature vectors of face patches, and third a majority vote decision based on these features.
To test our approach, new datasets including real and fake images are created. For producing fake images, StyleGAN and StyleGAN2 are trained by FFHQ images and also StarGAN and PGGAN are trained by CelebA images. The CelebA and FFHQ datasets are used as real images.
The proposed approach reaches higher results in early epochs than other methods and increases the SoTA results by 0.4\%-7.9\% in the different built data-sets. Also, we have shown in experimental results that weighing the patches may improve accuracy.}


\maketitle

\section{Introduction}\label{sec1}

The digital image forgery indicates the false and fraudulent alteration of an image content. Currently, people with a basic computer skill can manipulate images due to abundance of easy photo editing programs~\cite{c28}. The problem of counterfeiting is in particular sensitive with human images and there are more than one billion photos and videos uploaded each day. About 40 to 50 percent of them are manipulated, resulting in financial fraud and hoarding and even posing a security threat~\cite{c26}.
Nowadays, Deep learning and Generative Adversarial Networks (GANs) have enhanced the synthesis and manipulation of images, so that resulting images and videos are more realistic and difficult for human beings to detect even if they are fake. It is also impossible to detect whether they are real or fake using traditional forgery detection methods. 
	
\begin{figure}[htp]
		\begin{center}
			\includegraphics[width=.8\linewidth]{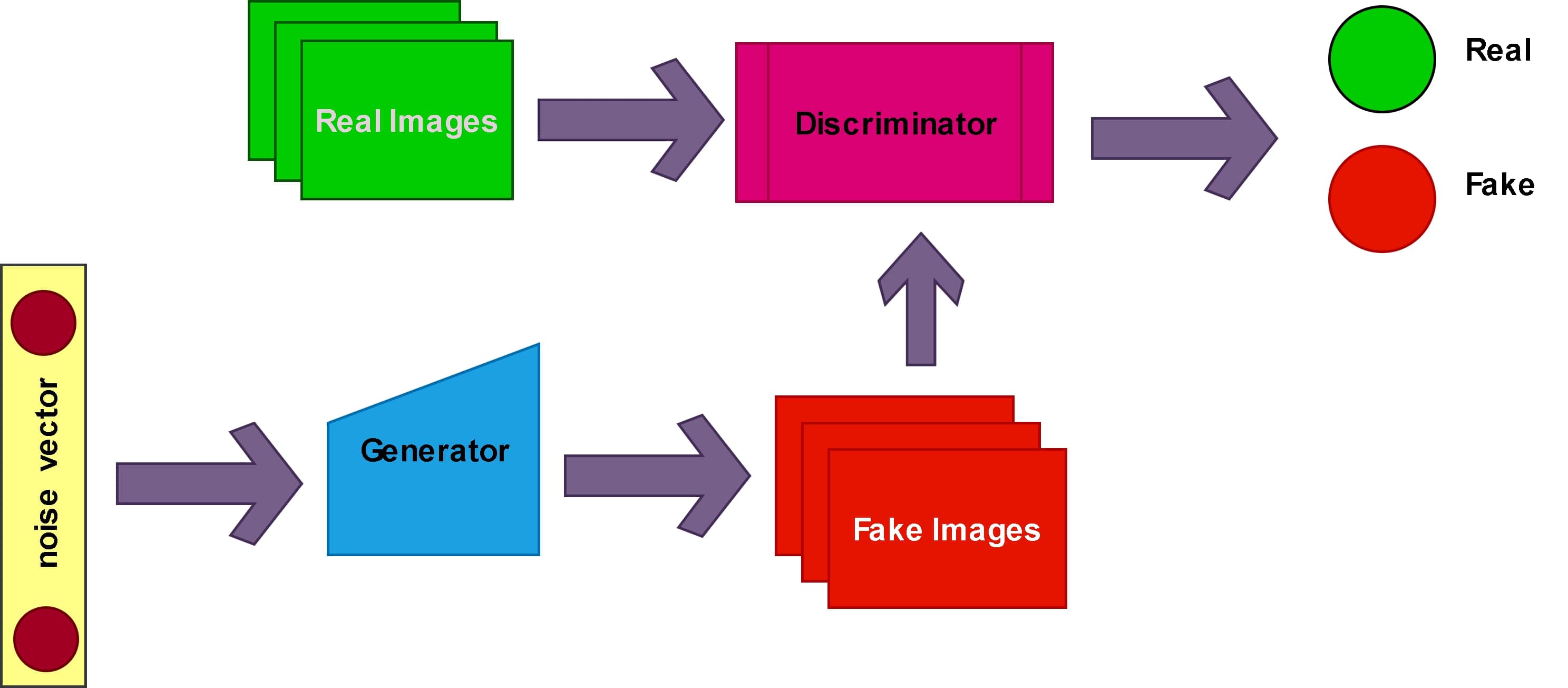}
			%\vspace{-15pt}
			\caption{Generative adversarial network structure}\label{gan}
		\end{center}
	\end{figure}

The use of GANs to create realistic images and videos has led to a technique called deepfake~\cite{c4}. GAN, introduced by Ian Goodfellow in 2014 and shown in Fig.\ref{gan}, consists of two neural networks: the generator and the discriminator~\cite{c4}.
	
\begin{figure}[htp]
	\begin{center}
		\includegraphics[width=.8\linewidth]{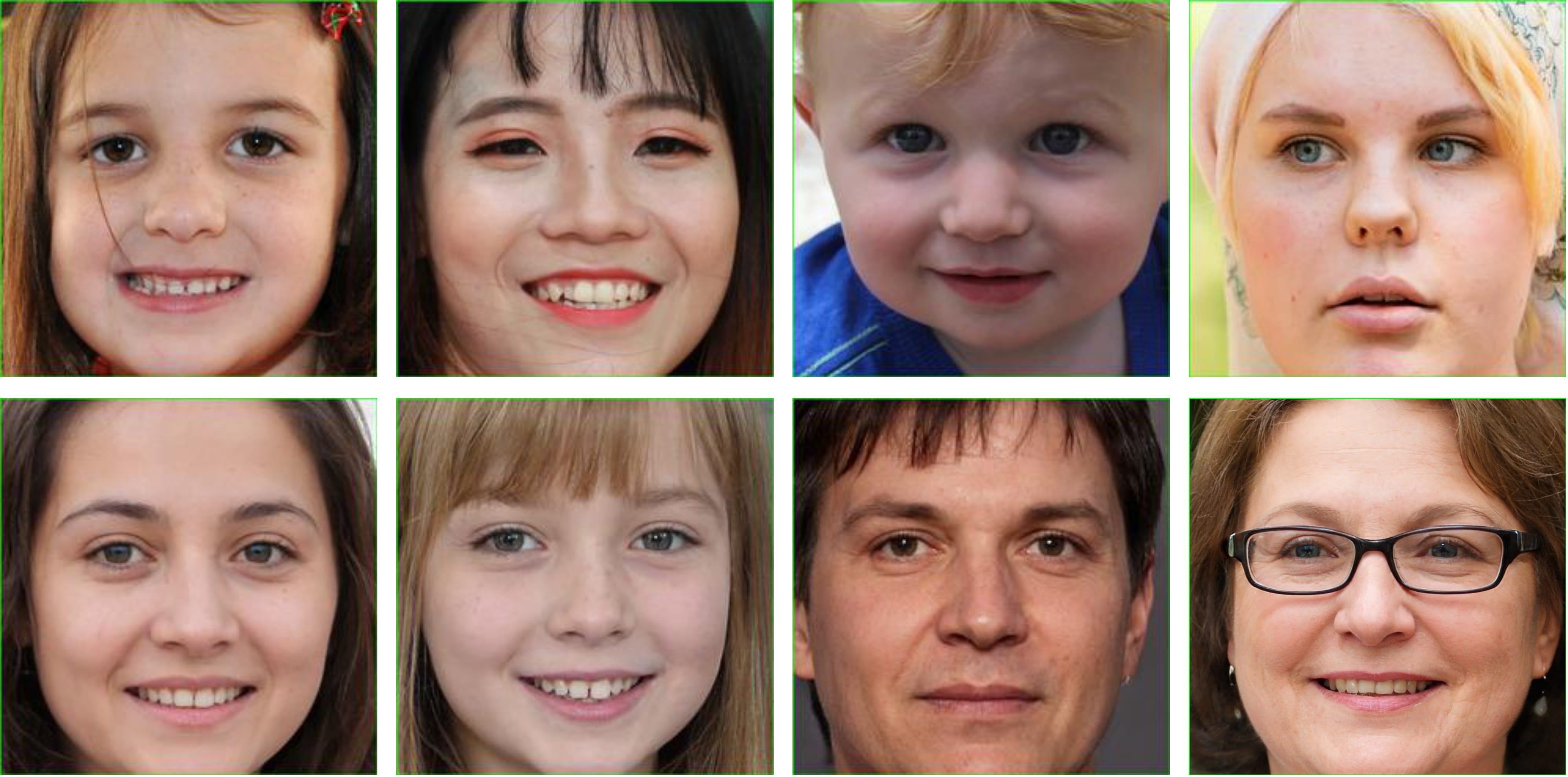}
		%\vspace{-15pt}
		\caption{The first row images are real images from FFHQ dataset and the second row images are images synthesized by StyleGAN2 with training by FFHQ dataset}\label{real_fake}
	\end{center}
\end{figure}

The manipulations created by GANs in face images, generally fall into four categories: full face synthesis, facial attribute manipulation, face identity swapping and expression swapping manipulation~\cite{c27}. 
Full face synthesis is realistic fake images generated from GANs. Facial attribute involves adding attributes to the face image such as hair and wrinkles and changing face image attributes such as hair color and skin color. 
Face identity swap refers to swapping faces of two people in a video. Expression swap refers to facial expressions transferred from one person to another. Fig.~\ref{real_fake} is real or a synthesize image?
	
In this paper, the proposed method is related to full face synthesis. The proposed method detects synthesized images, difficult for human eyes to detect. It is based on a patch-based proposal with a multi-path decision to detect real images from fake ones in the presence of occlusion. Contributions of this paper are:
	
	\begin{itemize}
		\item Despite other articles using a single method to make a decision, based on our investigation, we are first to employ different methods together not only to reach a decision but also to be assured of a final decision.
		
		\item The use of a multi-path decision based on deep learning to distinguish fake images from real ones leads to higher performance.
		
		\item We examine occlusion and two different splitting methods, the same size block and semantically face components in detecting fake images.
		
		\item Occlusion removal phase and weighing facial patches increase the accuracy of fake/real image detection.
		\end {itemize}
		
During this research, we utilize fake images generated by StyleGAN2 and StyleGAN, trained using FFHQ images, and others created by StarGAN and PGGAN, trained using CelebA images. We also use FFHQ and CelebA images as real images.
		
This paper is structured as follows. In the second part, we review the previous works done in the field of deepfake images detection. In third section, we describe the proposed method, Details about the data are provided in the fourth section, in the fifth section, we cover model evaluation, model training, experiments and results, the sixth section is dedicated to human studies, in the seventh section, we discuss and analyze further and finally, in the eighth section, the paper is summarized.	

\section{Related work}
There are two general methods for distinguishing GAN-generated images from real ones: image forensic methods and deep learning methods. Image forensics methods usually use statistical properties of image pixels and colors to identify whether they are real or fake, whereas deep learning methods treat the problem as a two-class classification. Li et al.~\cite{c5} and McCloskey~\cite{c7} used forensic imaging techniques and features of the GAN pipeline to distinguish real images from fake GAN images. The most common methods nowadays to distinguish fake GAN images from real ones, have been deep learning methods using color image features, GAN pipeline features, paired learning, face patches, texture, and attention mechanisms.
Li et al., presented a study in 2018~\cite{c5}, used statistical features of images and feature extraction using a co-occurrence matrix for fake detection. They analyzed the color components of the images and patterns of images in RGB, HSV and YCbCr color spaces. For better detection, they obtained the residual image using a high-pass filter, and after pre-processing, calculated the co-occurrence matrix for each component of the color space. Finally, the obtained feature vectors are concatenated and used to classify whether images were fake or real. 
McCloskey and Albright used color values and SVMs for fake detection in 2018 by examining the common features of GANs. They were able to detect fake images from real ones by using color and normalization features, and by looking at the frequency and relationships between noise and intensity values~\cite{c7}.
Yu et al.~\cite{c11} and Wang et al.~\cite{c13} conducted studies based on GAN features. Yu et al. defined fingerprints based on image features and GAN parameters in 2019. These GAN parameters are its network architecture, distribution of training data sets, loss function, optimization strategy, and hyper-parameter settings. Differences in each of these parameters result in a unique GAN fingerprint. For fingerprints of images, they examined that images, produced by one GAN, had common patterns with images produced by a similar GAN, and were different from images produced by other GANs. They considered this property as the fingerprint of images. They used an Attribution network to recognize real images and  GAN-generated images. They examined their proposed approach's robustness for detection of noisy images , blurred images, cropping, jpeg compression, and relighting, and found that it did not work appropriately for detection of noisy images  and jpeg compression~\cite{c11}.
Wang et al. in 2019 used neuron coverage for fake detection by calculating an optimal threshold for neuron activation in each layer and the number of neurons that their outputs exceeded the threshold in each layer. They calculated feature vectors and used a binary classifier with five fully connected layers. They investigated detection of blurred images, resized images, compressed images, and noisy images at different intensities and found that AUC decreased by less than 3.77\% in all 4 challenges with different intensities~\cite{c13}.
Additionally, the attention mechanism with a 2-class convolutional neural network classifier for feature mapping in 2020 was used by Deng et al.~\cite{c14}. Attention highlights areas in an image and therefore improves diagnosis. They used the manipulation appearance model and direct regression to estimate the attention map and used supervised learning, weakly supervised learning, and unsupervised learning to compute the loss function resulting from the attention map~\cite{c14}. 
Furthermore, the use of texture for diagnosis has been reported in some studies. Liu et al. examined the importance and effect of texture on network's ability to distinguish fake from real images in 2020. For analyzing image textures and recording long-term information about image textures, introduced the Gram-Net network, derived by adding 6 Gram blocks to the ResNet18 architecture. Each Gram block in the ResNet18 architecture contained one convolution layer, one Gram matrix~\cite{c35} layer to extract image texture properties, two convolution layers, batch normalization, Relu, and one pooling layer~\cite{c1}.
Some works \cite{c2,c3,c23} used facial patches for diagnosis. In 2018, Jain et al. introduced an architecture consisting of 5 convolutional layers and 2 fully connected layers, and a wide residual connection in order to add the output of the second layer to the fifth layer. 
A threshold-based approach and prediction based on threshold were used to classify images as real or fake. They first divided an image into $64\times64$ patches, predicted fake patches by their model, and calculated the threshold by dividing the number of fake patches to the total number of patches~\cite{c2}.
In 2020, Jane et al.~\cite{c3} introduced a three-level patch-based approach for fake detection.In the first level, they divided an image into $64\times64$ blocks, and obtained a vector for each image. They also trained a SVM with RBF kernel with these vectors to determine whether an image was real or altered. On the second level, weights of the previous layer were frozen, and if an image has been altered, the network detected that images have been retouched or generated by a GAN.  Finally, in the third level, when an image was produced by a GAN, weights of the previous level were frozen, and at this level, the type of GANs was detected. 
Bharati et al, in 2016, proposed considering four facial patches, including the right periocular, left periocular, nose, and mouth. They trained a Deep Boltzmann machine for each patch. The output features obtained from the SDBM for each patch were concatenated and fed to a SVM for classification~\cite{c23}. 
The color features of images were used by Nataraj et al. for fake detection by applying a co-occurence matrix to three color channels of the image. They obtained a $3\times256\times256$ tensor for each image, and then trained their proposed CNN for fake classification~\cite{c17}.
Zhuang et al. in 2020 used a two-step learning for diagnosis \cite{c15,c18}.
In 2019, Zhuang et al. used a coupled deep neural network and two-step learning approach to learn common fake features. Their CDNN was based on three residual units. The first residual unit consisted of two residual blocks with 96 channels. In the second and third residual units, there were respectively four and three residual blocks with 128 and 256 channels. They used triplet loss as their loss function~\cite{c15}
through the hierarchical feature representation and medium-level feature representation with the contrastive loss as loss function.
The hierarchical feature representation produced by a CNN architecture. The medium-level feature representation was used to find discriminative common fake features(CFFs) by their proposed CFFN. 
CFFN consisted of three dense units, including two, four, and three dense blocks.~\cite{c18}. 
Ciftc et al. introduced a diagnostic method based on biological signals in 2020, called FakeCatcher. They presented a new database called DeepFakes in their paper and claimed that FakeCatcher was robust to low resolution, motion, lighting, occlusion, and cosmetic artifacts~\cite{c36}.
		
Table \ref{related work} shows the related studies in brief.
		
	\begin{table}[htp]
		\begin{center}
			\caption{Summary of the related work}
			\label{related work}
			\renewcommand{\arraystretch}{1.5}
			\setlength{\extrarowheight}{20pt}
			\scalebox{0.5}{%
				\begin{tabular}{|@{}l | l| l|l @{}|}
					\toprule
					\;\;\;\;Refrence & Features & Classifications & Datasets \\ \midrule
					\;\;\;\;\shortstack{McCloskey et al. (2018) \\ ~\cite{c7}} & Using the color and common properties of the  GAN & SVM and VGG &  NIST MFC2018\\
					\hline
					\;\;\;\;\shortstack{Jain et al. (2018) \\ ~\cite{c2}} & \shortstack{Their proposed torsional neural network} & SVM and Threshold &  ND-IIITD, own(StarGAN)\\
					\hline
					\;\;\;\;\shortstack{Yu et al. (2019) \\ ~\cite{c11}} & the use of fingerprint images and GANs & CNN &  \shortstack{CelebA, LSUN, ProGAN, \\ SNGAN, CamerGAN, MMDGAN}\\
					\hline
					\;\;\;\;\shortstack{Zhuang et al. (2019) \\ ~\cite{c15}} & \shortstack{the proposed coupled deep neural network \\ with a two-step learning and triplet loss} & Fully connected layers &  \shortstack{CelebA, DCGAN, WGap, \\ WGAN-GP, LsGAN, PGGAN}\\
					\hline
					\;\;\;\;\shortstack{Zhuang et al. (2020) \\ ~\cite{c18}} & \shortstack{the proposed coupled deep neural network \\ with a two-step learning and constrastive loss} & Fully connected layers &  \shortstack{CelebA, DCGAN, WGap, \\ WGAN-GP, LsGAN, PGGAN}\\
					\hline
					\;\;\;\;\shortstack{Wang et al. (2019) \\ ~\cite{c13}} & \shortstack{Covering active neurons in CNNs} & \shortstack{Using a shallow neural  network \\ with  5 fully connected layers} &  \shortstack{CelebA,FFHQ \\ StyleGAN2,PGGAN}\\
					\hline
					\;\;\;\;\shortstack{Bharati et al. (2016) \\ ~\cite{c23}} & \shortstack{SBRM} & SVM &  CelebA, ND-IIITD\\
					\hline
					\;\;\;\;\shortstack{Jain et al. (2020) \\ ~\cite{c3}} & DAD-HCNN & SVM &  \shortstack{CMU Multi-PlE, ND-IIITD,\\ StarGAN, DCGAN, Context Encoder}\\
					\hline
					\;\;\;\;\shortstack{Liu et al. (2020) \\ ~\cite{c1}} & Image texture dependent & Gram-Net &   StarGAN, DCGAN, StyleGAN\\
					\hline
					\;\;\;\;\shortstack{Nataraj et al. (2019)\\ ~\cite{c17}} & Co-occurrence matrix on all three color channels of the image & CNN &   StarGAN, DCGAN, StyleGAN\\
					\hline
					\;\;\;\;\shortstack{Dang et al. (2020) \\ ~\cite{c15}} & Image property dependent & Fully connected layers &   \shortstack{DFFD, Faceforensics++, \\ CelebA, FFHQ}\\
					\hline
					\;\;\;\;\shortstack{Li et al. (2018)\\ ~\cite{c5}} & Color and Co-Occurence Matrix & SVM and LDA&   \shortstack{CelebA, LFW, PGGAN \\ WGAN-GP, DCGAN, DFC-VAE}\\
					\hline
                    \;\;\;\;\shortstack{Ciftc et al. (2020\\ ~\cite{c36}} & biological signals & CNN &   \shortstack{Face Forensics,Face Forensics++, UADFV, \\ Celeb-DF, own (Deep Fakes)}\\
                    \hline

			\end{tabular}}
		\end{center}
	\end{table}

\section{Proposed Method}
\begin{figure}[htp]
			\begin{center}
				\includegraphics[width=.8\linewidth]{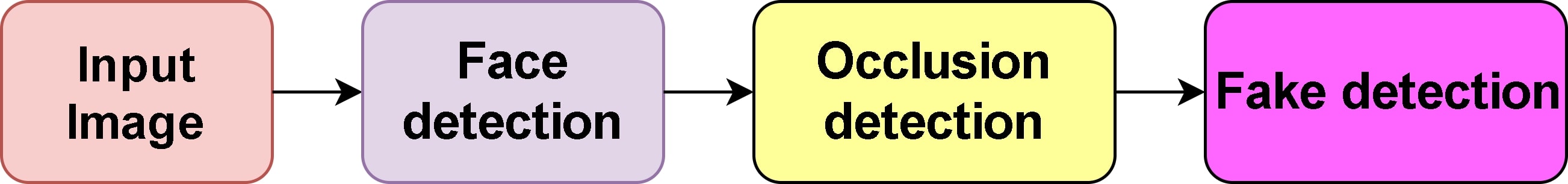}
				%\vspace{-15pt}
				\caption{The steps of the proposed model for distinguishing real images from images obtained by deepfake}\label{fig1}
			\end{center}
		\end{figure}

To distinguish between genuine and fake face-images, we consider a patch-based approach with the occlusion challenge as one of the challenges in Deepfakes. According to Fig.~\ref{fig1}, Our proposed approach generally includes face detection, occlusion detection, and fake detection.
		
First, we use the predominant face detectors such as Viola Joens and deep learning methods such as MTCNN to obtain the face of a person in the image. We then use one of the semantic segmentation methods\footnote{https://github.com/shaoanlu/face\_toolbox\_keras} from a pre-trained network to detect occluded areas. If it is polluted with occlusion, occluded areas are removed. Finally, we utilize a multi-path decision including three parallel diagnostic procedures based on the Gram-Net network to determine whether the image is real or fake.
		
In the first path of the diagnostic phase, we pass the entire face image to a fake detector model to specify whether it is real or not. The other two diagnostic procedures are based on patches, in which the face image is divided into a number of patches. 
		
The second diagnostic procedure involves dividing the face into patches, then obtaining a feature vector for each patch. By concatenating all the feature vectors from all the patches, we can detect whether the image is fake or real. 
By using the third diagnostic procedure, we can determine if each patch is real or fake. If the number of fake patches is greater than the number of real patches, we deem the image fake but otherwise real. 
Since each of the three diagnostic procedures determines whether an image is real or fake, if two approaches determine that the image is real and one approach determines whether it is fake, the image is considered real, otherwise it is determined to be fake. 
		
\begin{figure}[htp]
\begin{center}
		\includegraphics[width=.8\linewidth]{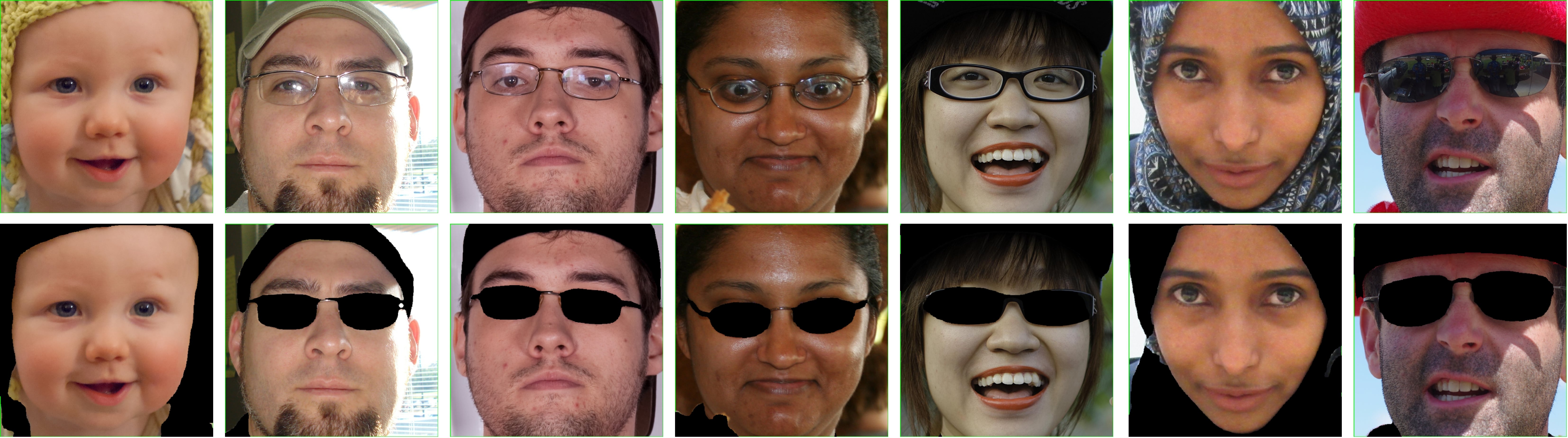}
		\caption{examples of face image occluded area detection }
		\label{occ}
\end{center}
\end{figure}

\begin{figure}[htp]			
			\includegraphics[width=\linewidth]{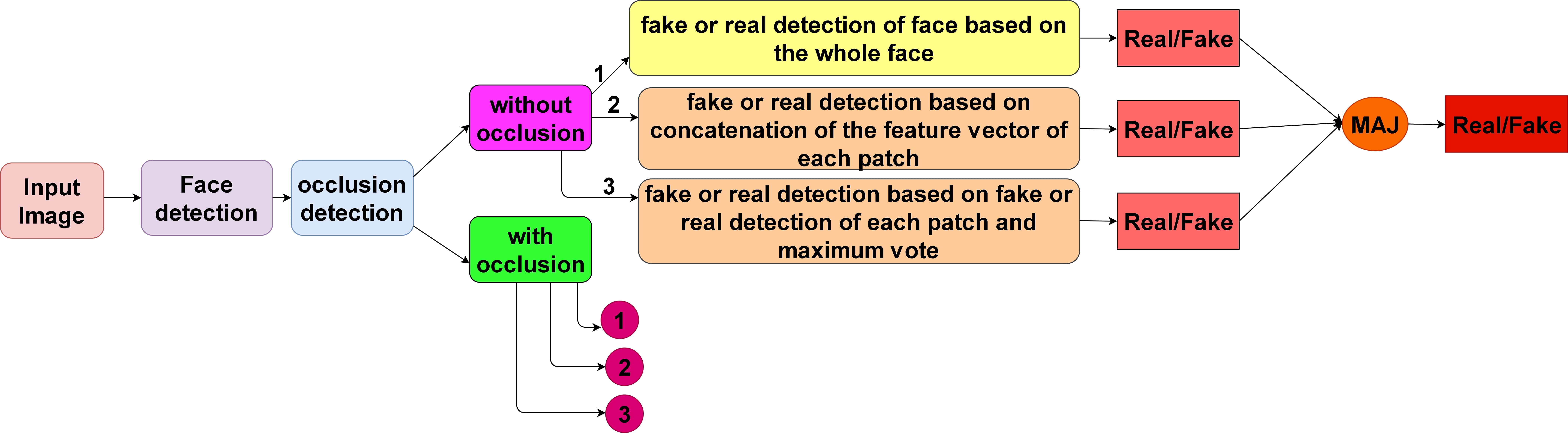}
			\caption{General diagram of the proposed approach}
			\label{fig2}
		\end{figure}

\begin{figure}[htp]
			\includegraphics[width=\linewidth]{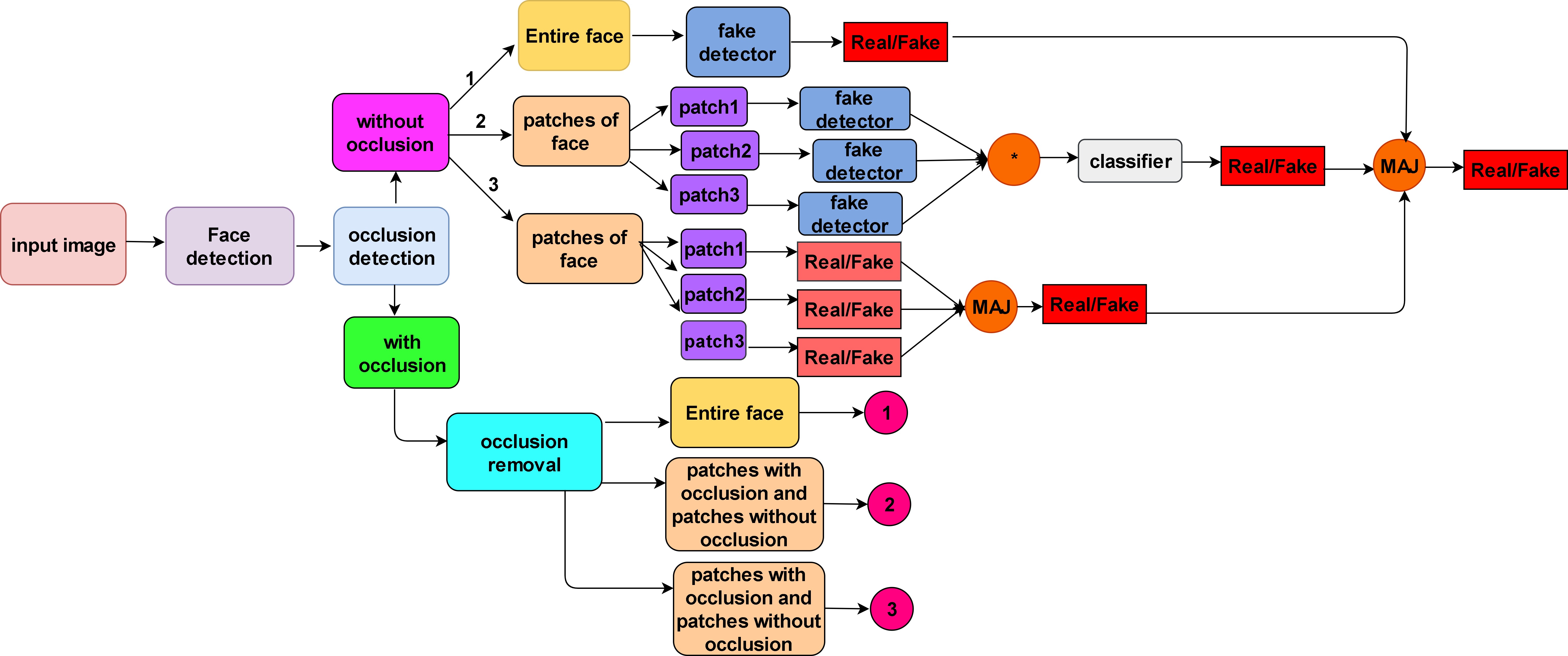}
			\caption{Fake Face Detector Flowchart Based on Combining Patch-Based Methods and Considering Occluded Images}
			\label{fig3}
		\end{figure}

For occluded images, we have the same trend, except that pixels of occluded areas are converted to zero. The proposed structure is shown in more detail in Fig.~\ref{fig2} and Fig.~\ref{fig3}. 
Occlusion detection block in the proposed structure detects occluded areas and removes pixels of occluded areas. Examples of this occlusion removal block are shown in Fig.~\ref{occ}.

\begin{figure}[htp]
			\includegraphics[width=\linewidth]{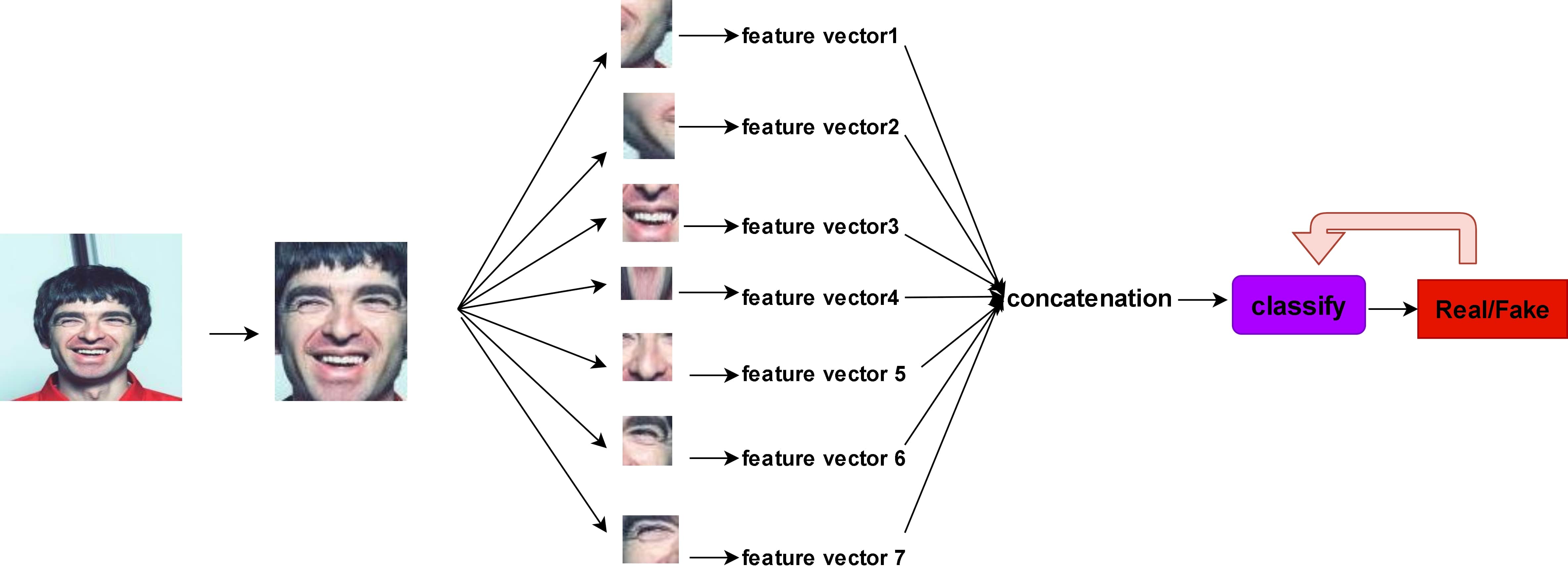}
			\caption{Training process of second diagnostic procedure for an image without occlusion}
			\label{fig4}
		\end{figure}

In the implementation of our proposed approach, we use seven patches including the right cheek, left cheek, mouth, nose, chin, right eye, and left eye based on the facial landmarks extracted from the images by dlib. In Fig.~\ref{fig4}, the training process is shown for the second diagnostic procedure. After dividing the face into patches, each patch is given to Gram-Net, and the feature vector from the layer before the last layer is obtained for each patch, which in our method has 704 dimensions and the concatenation of feature vectors is fed to a classifier to determine whether the image is real or fake and finally the cross entropy loss function is used to train the network.

\begin{figure}[htp]
	\centering
	\includegraphics[width=0.8\linewidth]{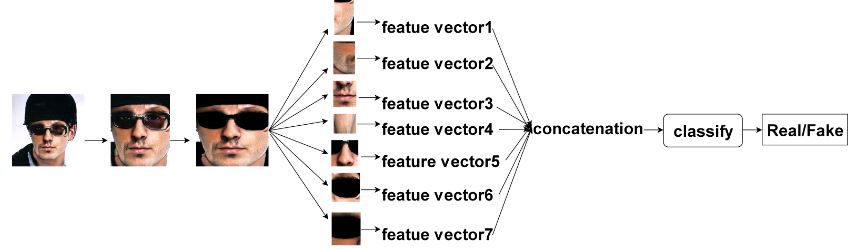}
	\caption{Second diagnostic procedure for an occluded face image}
	\label{fig5}
\end{figure}

\begin{figure}[htp]
	\centering
	\includegraphics[width=0.8\linewidth]{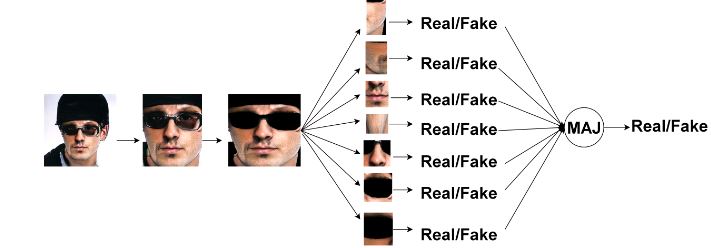}
	\caption{Third diagnostic procedure for an occluded face image}
	\label{fig6}
\end{figure}

In the second and third patch-based diagnostic approaches, we train the model with all facial patches obtained from the training set.
Fig.~\ref{fig5} and Fig.~\ref{fig6}  show the test phase for the second and third procedures for an occluded face image.		
		
\begin{figure}[htp]
			\centering
			\includegraphics[width=0.8\linewidth]{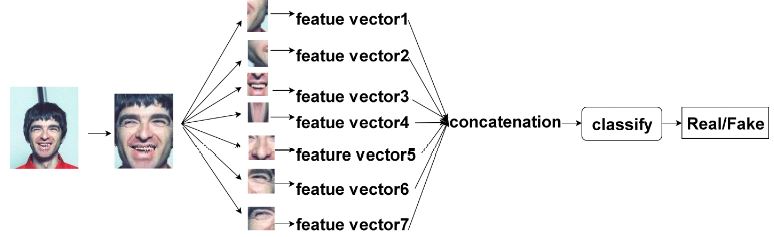}
			\caption{Second diagnostic procedure for a face image without occlusion}			
			\label{fig7}
		\end{figure}

\begin{figure}[htp]
			\centering
			\includegraphics[width=0.8\linewidth]{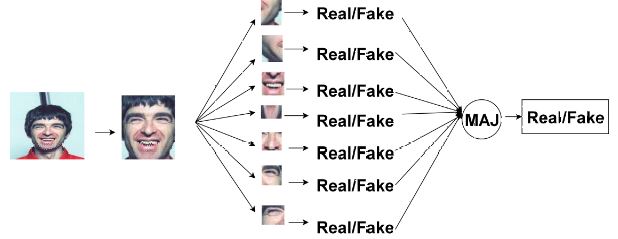}
			\caption{Third diagnostic procedure for an image without occlusion}
			\label{fig8}
		\end{figure}

Fig.~\ref{fig7} and Fig.~\ref{fig8}  show the test phase for the second and third procedures for a face image without occlusion.
		
%		\hfill\break				

\section{Datasets}
Each of our built datasets includes real and fake images. The fake images produced by StyleGAN~\cite{c31} and StyleGAN2~\cite{c33} networks, use FFHQ~\cite{c29} images for training. Moreover, those images produced by StarGAN~\cite{c32} and PGGAN~\cite{c34} use CelebA~\cite{c30} images as training. The CelebA and FFHQ datasets are used as real images. CelebA images include 202,599 face images of 10,177 unique celebrities with large changes in posture, background disorder and diversity. There are 52,000 high-quality PNG images in the Flickr-Faces-HQ Dataset (FFHQ). Each image is $512\times 512$ pixels and with significant differences in age, ethnicity, and background. It also provides good coverage of accessories, such as sunglasses and hats. Five built datasets used in this article are mentioned in Table \ref{table1} with the number of training, testing, and validation images for each data-set.

		\begin{table}[htp]
			\centering
			\caption{Details of Datasets. Second to four columns show the source of collecting images. Five to seven columns indicate the number of images for each experimental phase.}
			\label{table1}
			\renewcommand{\arraystretch}{1.2}
			\setlength{\extrarowheight}{10pt}
			\begin{tabular}{|c|c|c|c|c|c|c|}
				\hline				
				Data-set & Train & Test & Validation & Train & Test & Validation\\  \midrule
				
				1 & \shortstack{StyleGAN \\ CelebA \\ + \\ StarGAN \\ FFHQ} & \shortstack{StarGAN \\ FFHQ} & & \shortstack{4066 \\ 3873} &
				\shortstack{152 \\ 397} & \\
				\hline
				2 & \shortstack{StyleGAN \\ CelebA} & \shortstack{StyleGAN \\ CelebA} & \shortstack{StyleGAN \\ CelebA} & \shortstack{3963 \\ 3681} & \shortstack{1546 \\ 1776} & \shortstack{671 \\ 843} \\
				\hline
				3 & \shortstack{StyleGAN2 \\ FFHQ} & \shortstack{StyleGAN2 \\ FFHQ} & \shortstack{StyleGAN2 \\ FFHQ} & \shortstack{4671 \\ 4802} & \shortstack{3005 \\ 2162} & \shortstack{1120 \\ 940} \\
                \hline
				4 & \shortstack{StyleGAN2 \\ FFHQ} & \shortstack{StyleGAN2 \\ FFHQ} & \shortstack{StyleGAN2 \\ FFHQ} & \shortstack{148 \\ 139} & \shortstack{100 \\ 90} & \shortstack{116 \\ 107} \\ \bottomrule
				\hline
                				
				5 & \shortstack{PGGAN \\ CelebA} & \shortstack{PGGAN \\ CelebA} & \shortstack{PGGAN \\ CelebA} & \shortstack{166 \\ 198} & \shortstack{152 \\ 199} & \shortstack{260 \\ 216} \\ 
				\hline
				
			\end{tabular}
			
		\end{table}
		
\section{Expriments and Results}
\label{results}
\subsection{Examining effects of two divisions of face images to patches}
There are two general ways to divide an image into a number of patches. One is to divide it  into a semantically defined number of patches. The second method is to divide it into a number of the same size blocks. At first, we examine the results of these two different methods on two datasets and use the method with the highest outcome in other datasets. Regarding the block division, we resize images to $256\times256$, then divide them into $64\times64$ and $64\times 128$ blocks. The semantic division is the dividing of the face into meaningful patches of the right eye, left eye, right cheek, left cheek, mouth, nose, and chin. 
Table \ref{table2} shows the results of the two types of block division methods and the semantic patch method for the first dataset. Table \ref{table3} presents the results of the two types of block division methods and the semantic patch method for the fourth dataset. 
		
As shown in Table \ref{table2}, block division into $64\times128$ blocks for the first dataset has an accuracy of 98.3\%, the lowest accuracy among the two division approaches. The block division of $64\times64$ is 100\% accurate. Additionally, Table \ref{table3} depicts that the division into $64\times128$ blocks for the fourth data set has 84.7\% accuracy, which is the lowest accuracy compared to the two divisions: the semantic division approach with an accuracy of 90.5\% and a division into $64\times64$ blocks with an accuracy of 90\%. 
The $64\times64$ block division is almost as accurate as the semantic division. We have  conducted our experiments with the semantic division for two reasons. 
First, it causes a reduction in temporal complexity because in the same size division, each image is divided into 16 patches of $64\times64$ blocks. Whereas in the semantic division, each image is divided into seven meaningful patches. The use of the semantic division will halve the complexity and train the network more quickly due to the smaller number of patches. Another reason is that inference in the semantic division of the image into seven patches can be drawn easier than the block division of 16 patches because the number of patches is odd. Whereas in the same size block division of 16 patches, if equal real and fake patches are detected, e.g 8 real and 8 fake patches, the network based on the entire face will declare a class as the final prediction and two procedures based on patches can not help the decision.
		
		\begin{table}[htp]
			\centering
			\caption{The results of the different patch types for the first dataset}
			\label{table2}
			\renewcommand{\arraystretch}{1.2}
			\setlength{\extrarowheight}{10pt}
			\begin{tabular}{|@{}l | l | l | l | l| l @{}|}
				\toprule
				& accuracy & precision & recall & f-score & \shortstack{confusion \\ matrix} \\ \midrule
				\shortstack{Block division \\ $64\times128$ } & 98.3 & 94.9 & 99.34 & 97.06 & \shortstack{$[\! \;151,1\; \!]$, \\ $[ \!\;8,389\; \! ]$}  \\
				\hline
				\shortstack{Block division \\ $64\times64$ } & 100 & 100 & 100 & 100 & \shortstack{$[\! \;152,0\; \!]$, \\ $[ \!\;0,397\; \! ]$}  \\
				\hline
				Semantic division & 100 & 100 & 100 & 100 & \shortstack{$[\! \;152,0\; \!]$, \\ $[ \!\;0,397\; \! ]$}  \\
				\hline

			\end{tabular}
		\end{table}
		
		\begin{table}[htp]
			\centering
			\caption{The results of the different patch types for the fourth dataset}
			\label{table3}
			\renewcommand{\arraystretch}{1.2}
			\setlength{\extrarowheight}{10pt}
			\begin{tabular}{|@{}l | l | l | l | l| l @{}|}
				\toprule
				& accuracy & precision & recall & f-score & \shortstack{confusion \\ matrix} \\ \midrule
				\shortstack{Block division \\ $64\times128$ } & 84.7 & 86.50 & 84.00 & 85.23 & \shortstack{$[\! \;84,16\; \!]$, \\ $[ \!\;13,77\; \! ]$}  \\
				\hline
				\shortstack{Block division \\ $64\times64$ }  & 90 & 92.6 & 88 & 90.24 & \shortstack{$[\! \;88,12\; \!]$, \\ $[ \!\;7,83\; \! ]$}  \\
				\hline
				Semantic division  & 90.5 & 91.83 & 90 & 90.9 & \shortstack{$[\! \;90,10\; \!]$, \\ $[ \!\;8,82\; \! ]$}  \\
				\hline
			\end{tabular}
		\end{table}		

\subsection{Model training}
In this paper, the Pytorch framework is used to implement the proposed method. To train the models, we used the SGD Optimizer and cross-entropy loss function. In all experiments, the learning rate, number of epochs, and batch size are set to 0.002, 30 and 15 respectively.
		
\subsection{Evaluation Metrics}
\begin{figure}[htp]
			\centering
			
			\includegraphics[width=0.4\linewidth]{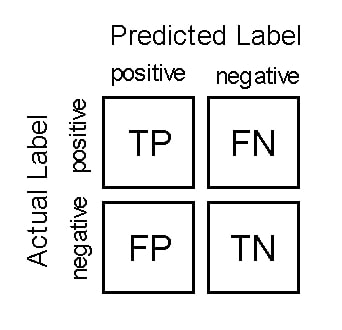}
			\caption{Confusion matrix}
			\label{confusion_matrix}
		\end{figure}

The proposed model is evaluated by using four metrics, including accuracy (ACC), precision (PR), recall (RE) and F-measure given in Equations 1 to 4. We also use the confusion matrix as in Fig.~\ref{confusion_matrix} where the entry TP shows that the actual and predicted label are positive. The entry TN also represents that the actual and predicted label are negative. The entry FN is when the actual label is positive and the predicted label is negative. The entry FP is when the actual label is negative and the predicted label is positive. Assuming that the negative class represents fake images while the positive class represents real images. TN represents the number of fake images that has been detected as fake. FP represents the number of fake images that has been detected as real. FN represents the number of real images that has been detected as fake. Finally, TP represents the number of real images that has been detected as real. In this paper, the macro version of precision, recall and F-measure are calculated and reported.
		
		\begin{equation}
			ACC=\frac{TP+TN}{TP+FP+TN+FN}
		\end{equation} 
		
		\begin{equation}
			PR=\frac{TP}{TP+FP}
		\end{equation}
		\begin{equation}
			RE=\frac{TP}{TP+FN}
		\end{equation}
		\begin{equation}
			F\_measure=\frac{2\times PR\times RE}{PR+RE}
		\end{equation}
		
\subsection{Results}
\subsubsection{First dataset results}
%\hfill\break
The first data-set contains the fake images produced by StyleGAN and StarGAN, and the real images from CelebA and FFHQ. In Table~\ref{table4}, the proposed approach for this dataset achieves 100\% accuracy, which is better than~\cite{c1} and ~\cite{c3}. Lin et al.~\cite{c1} and Jain et al.~\cite{c3} have 98.3\% and 79.4\% accuracy, respectively. From the confusion matrix in Table \ref{table4}, it can be seen that 9 fake images are detected as real images in the Lin et al.~\cite{c1} and 113 real images as fake in Jain et al.~\cite{c3}.

		\begin{table}[htp]
			\centering
			\caption{Results of the first dataset}
			\label{table4}
			\renewcommand{\arraystretch}{1.2}
			\setlength{\extrarowheight}{10pt}
			\begin{tabular}{|@{}c | c | c | c | c | c @{}|}
				\toprule
				Method & Accuracy & Precision & Recall & F-Score & \shortstack{Confusion \\ Matrix} \\ \midrule
				\;\;\;\;\;\;Lin et al.~\cite{c1} & 98.3 & 97.20 & 98.86 & 97.98 & \shortstack{$[\! \;152,0\; \!]$, \\ $[ \!\;9,388\; \! ]$}  \\
				\hline
				
				\;\;\;\;\;\;Jain et al.~\cite{c3} & 79.4 & 88.9 & 62.8 & 64.1 & \shortstack{$[\! \;39,113\; \!]$, \\ $[ \!\;0,397\; \! ]$}\;  \\
				\hline
				\shortstack{\;\;\;The proposed \\method} & 100 & 100 & 100 & 100 & \shortstack{$[\! \;152,0\; \!]$, \\ $[ \!\;0,397\; \! ]$}\;  \\ 
				
				\hline

			\end{tabular}
		\end{table}
		
\subsubsection{Second dataset results}
\begin{figure}[htp]
			\centering
			\includegraphics[width=0.8\linewidth]{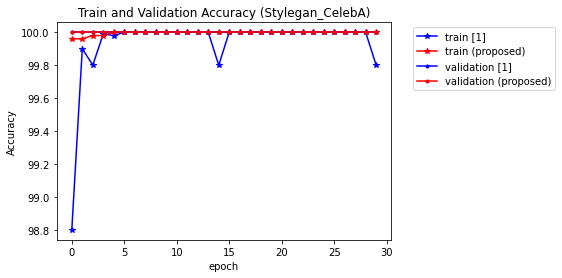}
			\caption{This diagram compares Lin et al.~\cite{c1} with the proposed approach for the training and validation phases for the second dataset in 30 epochs}
			\label{acc_chart_second}
		\end{figure}

		%\hfill\break
The second dataset contains fake images produced by StyleGAN and real images from CelebA. In Table \ref{table5}, our proposed approach and Lin et al.\cite{c1} and Jain et al.\cite{c3} have 100\% accuracy, whereas Jain et al.'s another paper~\cite{c2} has the lowest accuracy using the threshold and the SVM method in its classification. The accuracy diagram of the training and validation phases with 30 epochs for Lin et al.\cite{c1} and the proposed approach for the second dataset is shown in Fig.~\ref{acc_chart_second}.
		
	\begin{table}[htp]
			\centering
			\caption{Results of the second dataset}
			\label{table5}
			\renewcommand{\arraystretch}{1.2}
			\setlength{\extrarowheight}{10pt}
			\begin{tabular}{|@{}c | c | c | c | c | c @{}|}
				\toprule
				Method & Accuracy & Precision & Recall & F-Score & \shortstack{Confusion \\ Matrix} \\
				
				\midrule
				\shortstack{~Lin et al.~\cite{c1}} & 100 & 100 & 100 & 100 & \shortstack{$[\! \;1546,0\; \!]$, \\ $[ \!\;0,1776\; \! ]$}  \\
				
				\hline
				\shortstack{~Jain et al.~\cite{c2} \\~threshold-based\\classification} & 59 & 76.8 & 62.3 & 54 & \shortstack{$[\! \;1546,0\; \!]$, \\ $[ \!\;1337,439\; \! ]$}  \\
				\hline
				
				\shortstack{~Jain et al.~\cite{c2} \\~SVM\\classification} & 91.4 & 91.4 & 91.6 & 91.4 & \shortstack{$[\! \;1459,87\; \!]$, \\ $[ \!\;196,1580\; \! ]$}\;  \\
				
				\hline
				\shortstack{~Jain et al.~\cite{c3}} & 100 & 100 & 100 & 100 & \shortstack{$[\! \;1546,0\; \!]$, \\ $[ \!\;0,1776\; \! ]$}\;  \\
				
				\hline
				\shortstack{~Our proposed \\ method} & 100 & 100 & 100 & 100 & \shortstack{$[\! \;1546,0\; \!]$, \\ $[ \!\;0,1776\; \! ]$}\;  \\ 
				
				\hline

			\end{tabular}
		\end{table}
		
\begin{figure}[htp]
			\centering
			\includegraphics[width=0.8\linewidth]{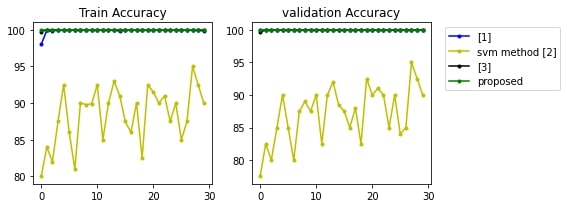}
			\caption{This Diagram compares the proposed approach with~\cite{c1,c2, c3} for the training and validation phases for the second dataset in 30 epochs.}
			\label{acc_chart_second1}
		\end{figure}

The accuracy charts for the training and validation phases of~\cite{c1,c2, c3} and the proposed method with 30 epochs for the second dataset are shown in Fig.~\ref{acc_chart_second1}.
		
\subsubsection{Third dataset results}
\begin{figure}[htp]
			\centering
			\includegraphics[width=0.8\linewidth]{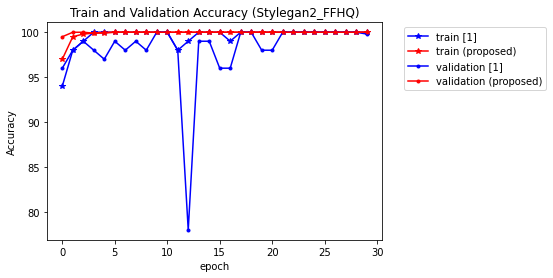}
			\caption{This diagram compares Lin et al.~\cite{c1} with the proposed approach for the training and validation phases of the third dataset in 30 epochs.}
			\label{acc_chart_third}
		\end{figure}

The third dataset contains fake images created by StyleGAN2 and real images from FFHQ.
On the basis of the Table~\ref{table6} results, Lin et al.'s study~\cite{c1} has an accuracy of 99.3\% and misidentified 32 images, whereas the proposed approach has an accuracy of 99.7\%, and  wrongly identify 14 images. The reference~\cite{c3} has an accuracy of 94.7\%, and reference~\cite{c2} using the threshold approach has the lowest accuracy.
The results of the third data set are presented in Table~\ref{table6}. The accuracy diagram of Lin et al.~\cite{c1} and the proposed approach for the training and validation phases of the third dataset with 30 epochs is shown in Fig.~\ref{acc_chart_third}.
		
				\begin{table}[htp]
			\centering
			\caption{Results of the third dataset}
			\label{table6}
			\renewcommand{\arraystretch}{1.2}
			\setlength{\extrarowheight}{10pt}
			\begin{tabular}{|@{}c | l | l | l | l | c @{}|}
				\toprule
				Method & Accuracy & Precision & Recall & F-Score & \shortstack{Confusion \\ Matrix} \\
				\midrule
				\shortstack{~Lin et al.~\cite{c1}} & 99.3 & 99.355 & 99.37 & 99.359 & \shortstack{$[\! \;2988,17\; \!]$, \\ $[ \!\;15,2147\; \! ]$}  \\
				\hline
				\shortstack{~Jain et al.~\cite{c2}\\~threshold-based\\classification} & 61.6 & 67.7 & 54.7 & 48 & \shortstack{$[\! \;2914,91\; \!]$, \\ $[ \!\;1890,272\; \! ]$}  \\
				\hline
				\shortstack{~Jain et al.~\cite{c2}\\~SVM\\classification} & 73.5 & 74.1 & 70.7 & 71.2 & \shortstack{$[\! \;2632,373\; \!]$, \\ $[ \!\;995,1167\; \! ]$}\;  \\
				\hline
				\shortstack{~Jain et al.~\cite{c3}} & 94.7 & 94.6 & 94.5 & 94.5 & \shortstack{$[\! \;2873,132\; \!]$, \\ $[ \!\;140,2022\; \! ]$}\;  \\
				\hline
				\shortstack{The proposed \\ method} & 99.7 & 99.67 & 99.76 & 99.71 & \shortstack{$[\! \;2991,14\; \!]$, \\ $[ \!\;0,2162\; \! ]$}\;  \\ 
				\hline
			\end{tabular}
		\end{table}
		
\begin{figure}[htp]
			\centering
			\includegraphics[width=0.8\linewidth]{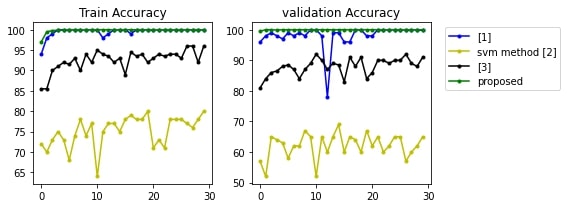}
			\caption{These charts display the comparison of the proposed approach with~\cite{c1,c2, c3} for the training and validation phases of the third dataset with 30 epochs.}
			\label{acc_chart_third1}
		\end{figure}

The accuracy charts of~\cite{c1,c2, c3} and the proposed approach for the training and validation phases of the third dataset with 30 epochs are shown in Fig.~\ref{acc_chart_third1}.
		
\subsubsection{Fourth Dataset Results}
\begin{figure}[htp]
			\centering
			\includegraphics[width=0.8\linewidth]{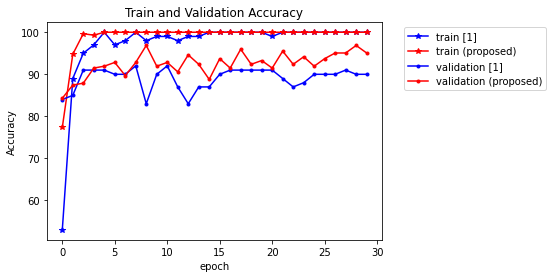}
			\caption{The diagram compares Lin et al.~\cite{c1} with the proposed approach for the training and validation dataset for the fourth dataset in 30 epochs}
			\label{acc_chart_fifth}
		\end{figure}

In the fourth dataset, fake images produced by StyleGAN2 and real images from FFHQ are used for the training, testing, and validation sets.
The proposed approach with an accuracy of 90.5\% has the highest accuracy in this dataset. The references, Jain et al.'s papers~\cite{c2,c3}, have an accuracy of 72.26\% and the reference~\cite{c1} has an accuracy of 82.6\%. The accuracy diagram of references~\cite{c1} and the proposed approach for the training and validation phases with 30 epochs for the fourth dataset is shown in Fig.~\ref{acc_chart_fifth}. The results of the fourth dataset are presented in Table~\ref{table8}.
		
\begin{figure}[htp]
			\centering
			\includegraphics[width=\linewidth]{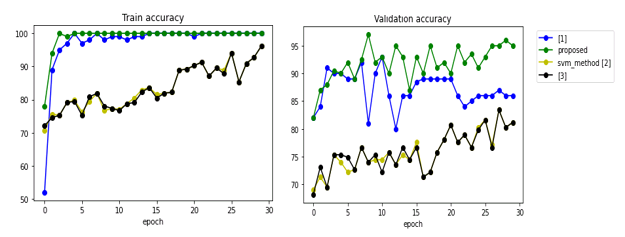}
			\caption{The diagram compares the proposed approach with~\cite{c1,c2, c3} for the training and validation phases of the fourth dataset with 30 epochs.}
			\label{acc_chart_fifth1}
		\end{figure}

		The accuracy charts of~\cite{c1,c2, c3} and the proposed approach for the training and validation phases with 30 epochs for the fourth dataset are shown in Fig.~\ref{acc_chart_fifth1}.
		
	\begin{table}[htp]
			\centering
			\caption{Results of the fourth dataset}
			\label{table8}
			\renewcommand{\arraystretch}{1.2}
			\setlength{\extrarowheight}{10pt}
			\begin{tabular}{|@{} c | l | l | l | l | c @{}|}
				\toprule
				Method & Accuracy & Precision & Recall & F-Score & \shortstack{Confusion \\ Matrix} \\
				
				\midrule
				\shortstack{~Lin et al.~\cite{c1}} & 82.6 & 85.12 & 83.33 & 82.48 & \shortstack{$[\! \;70,30\; \!]$, \\ $[ \!\;3,87\; \! ]$}  \\
				
				\hline
				\shortstack{~Jain et al.~\cite{c2} \\~with threshold\\classification} & 77.36 & 78.62 & 77.88 & 77.29 & \shortstack{$[\! \;68,32\; \!]$, \\ $[ \!\;11,79\; \! ]$}  \\
				
				\hline
				\shortstack{~\shortstack{Jain et al.~\cite{c2}}\\ wtih SVM \\classification} &75.26 & 78.08 & 76.05 & 74.95 & \shortstack{$[\! \;61,39\; \!]$, \\ $[ \!\;8,82\; \! ]$}\;  \\
				
				\hline
				\shortstack{~Jain et al.~\cite{c3}} & 75.26 & 78.08 & 76.05 & 74.95& \shortstack{$[\! \;61,39\; \!]$, \\ $[ \!\;8,82\; \! ]$}\;  \\
				
				\hline
				\shortstack{~The proposed \\ method} & 90.5 & 90.48 & 90.55 & 90.72 & \shortstack{$[\! \;90,10\; \!]$, \\ $[ \!\;8,82\; \! ]$}\;  \\ 
				
				\hline

			\end{tabular}
		\end{table}

\subsubsection{Fifth Dataset Results}
\begin{figure}[htp]
			\centering
			\includegraphics[width=0.8\linewidth]{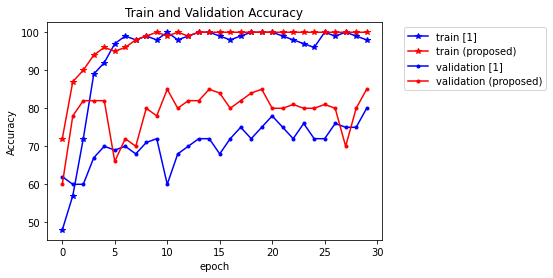}
			\caption{Th diagram compares Lin et al.~\cite{c1} with the proposed approach for the training and validation phases of the fifth dataset in 30 epochs.}
			\label{acc_chart_fourth}
		\end{figure}

In the fifth dataset, fake images produced by PGGAN and real images from FFHQ are used for the training, testing and validation sets.
In this dataset, the proposed approach with an accuracy of 84.9\% has the highest accuracy, while Jain et al.'s papers~\cite{c2},\cite{c3} have an accuracy of 55.5\% and Lin et al.~\cite{c1} have an accuracy of 80.3\%. The accuracy diagram of Lin et al.\cite{c1} and the proposed approach for the training and validation phases of the fifth dataset with 30 epochs is shown in Fig.~\ref{acc_chart_fourth}. The results of the fifth dataset are presented in Table~\ref{table7}. Due to the small volume of fake and real images in the training set for the fourth and fifth dataset, the Jain et al.'s papers~\cite{c2,c3} showed poor results. Moreover, we have attempted to improve their results by finding an optimal threshold within the 1-16 range, and then we normalize the data based on our datasets to ensure that Jain et al.'s papers~\cite{c2, c3}, have the highest level of accuracy, precision, and recall.

			\begin{table}[htp]
			\centering
			\caption{Result of the fifth dataset}
			\label{table7}
			\renewcommand{\arraystretch}{1.2}
			\setlength{\extrarowheight}{10pt}
			\begin{tabular}{|@{}c | l | l | l | l | c @{}|}
				\toprule
				Method & Accuracy & Precision & Recall & F-Score & \shortstack{Confusion \\ Matrix} \\
				
				\midrule
				\shortstack{~Lin et al.~\cite{c1}} & 80.3 & 76.79 & 81.26 & 80.3 & \shortstack{$[\! \;134,18\; \!]$, \\ $[ \!\;51,148\; \! ]$}  \\
				
				\hline
				\shortstack{~Jain et al.~\cite{c2}\\~threshold\\ classification} & 56.6 & 28.3 & 50 & 36.1 & \shortstack{$[\! \;0,152\; \!]$, \\ $[ \!\;0,199\; \! ]$}  \\
				
				\hline
				\shortstack{~Jain et al.~\cite{c2}\\~SVM\\classification} &56.6 & 28.3 & 50 & 36.1 & \shortstack{$[\! \;0,152\; \!]$, \\ $[ \!\;0,199\; \! ]$}\;  \\
				\hline
				\shortstack{~Jain et al.~\cite{c3}} & 56.6 & 28.3 & 50 & 36.1 & \shortstack{$[\! \;0,152\; \!]$, \\ $[ \!\;0,199\; \! ]$}\;  \\
				
				\hline
				\shortstack{~The proposed \\ method} & 84.9 & 85.04 & 85.66 & 84.84 & \shortstack{$[\! \;139,13\; \!]$, \\ $[ \!\;40,159\; \! ]$}\;  \\ 
				
				\hline

			\end{tabular}
		\end{table}

\begin{figure}[htp]
			\centering
			\includegraphics[width=0.8\linewidth]{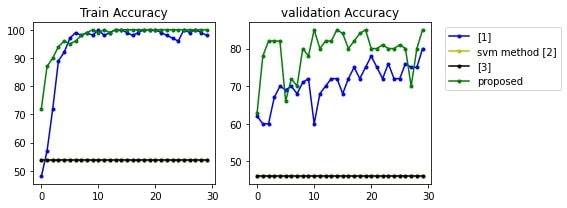}
			\caption{The diagram compares the proposed approach with~\cite{c1,c2, c3} for the training and validation phases of the fifth dataset in 30 epochs.}
			\label{acc_chart_fourth1}
		\end{figure}

The accuracy charts of~\cite{c1,c2,c3} and the proposed approach for the training and validation phased of the fifth dataset with 30 epochs are shown in Fig.~\ref{acc_chart_fourth1}.
		
\section{Human Study}
The fourth dataset includes 100 fake images produced by StyleGAN2 which is trained with the FFHQ dataset and 90 real images from the FFHQ dataset.
We use this dataset to observe the average accuracy of human diagnoses. During this study, we give these images to 20 participants and ask them to predict the labels without a time constraint and then we determine their detection rates based on the actual labels.
As Fig.~\ref{real_fake} shows, it can be difficult for human beings to distinguish between real and fake images since they are very realistic.
With this dataset, the highest diagnostic accuracy of these 20 people is 57.36\%, the lowest is 0\% and the average is 43.41\%, whereas our proposed method has an accuracy of 90.5\%. Table~\ref{human study} displays the accuracy of 20 predictions made by participants.
		
		\begin{table}[htp]
			\centering
			\caption{Results of human study for the fourth dataset}
			\label{human study}
			\renewcommand{\arraystretch}{1}
			\setlength{\extrarowheight}{10pt}
			\begin{tabular}{|@{} c | c | c | c @{}|}
				\toprule
				Participant & Accuracy & Participant & Accuracy \\ \midrule
				1           &44.37\%   & 11 & 47.25\% \\
				\hline
				2           &42.10\%  & 12 & 47.89\% \\
				\hline
				3           &45.26\%  & 13 & 44.50\% \\
				\hline
				4           &0.0\%    & 14 & 39\% \\
				\hline
				5           &57.36\%   & 15 & 45.15\% \\
				\hline
				6           &47.36\%  & 16 & 42.75\% \\
				\hline
				7           &43.15\%  & 17 & 48.27\% \\
				\hline
				8           &48.94\%   & 18 & 44.35\% \\
				\hline
				9           &47.25\%  & 19 & 42.15\% \\
				\hline
				10          &48.24\%  & 20 & 43.44\% \\
				\hline
			\end{tabular}
		\end{table}
		
\section{Discussion}
\subsection{Assessing the effect of occlusion rate on the accuracy of the proposed method}
In general, as shown by the results of Section~\ref{results}, the proposed approach is more accurate than references~\cite{c1,c2, c3} for the datasets given in Table~\ref{table1}. Based on the results, our proposed approach achieves an accuracy of 100\% for the first and second datasets, 99.7\% accuracy for the third dataset, 90.5\% accuracy for the fourth dataset, which is 7.9\% better than Lin et al. \cite{c1} which is the best refrence and 84.9\% accuracy for the fifth dataset, which is 4.6\% better than Lin et al.~\cite{c1} which is the best refrence, and 
In this section, we will examine the performance of the proposed approach to the occlusion challenge in more detail by defining the occlusion ratio which is the number of occluded images to all images within a dataset. 
Furthermore, we compare the performance of our model based on the occlusion ratio of each dataset with other studies to examine whether occlusion removal improves the performance or not. 
In Table~\ref{table10}, the occlusion rates in the experimental set for the third, fourth and fifth datasets are 0.387, 0.81 and 0.370 respectively. In the fourth and fifth datasets, in which the proposed method has the most improvement over Lin et al. \cite{c1} which is the best refrence, 81\%  and 37\% of the test images are occluded, respectively. Therefore, as much as the occlusion ratio increases, a further improvement of our proposed approach in comparison with the other studies~\cite{c1,c2, c3} happens. We can conclude that the occlusion removal phase is effective in improving the performance of the fake detection.
		
		\begin{table}[htp]
			\centering
			\caption{The occlusion ratio of all built datasets}
			\label{table10}
			\renewcommand{\arraystretch}{1}
			\setlength{\extrarowheight}{10pt}
			\begin{tabular}{|@{}c | c| c|c @{}|}
				\toprule
				~Dataset & train & test & validation~\\
				\midrule
				\hspace{2mm}1 & \hspace{2mm}0.09 & \hspace{2mm}0.16 & -\\
				\hline
				\hspace{2mm}2 & \hspace{2mm}0.08 & \hspace{2mm}0.19 &  \hspace{2mm}0.14 \\
				\hline
				\hspace{2mm}3 & \hspace{2mm}0.203 & \hspace{2mm}0.387 & \hspace{2mm}0.55 \\
				\hline
				\hspace{2mm}4 & \hspace{2mm}0.202 & \hspace{2mm}0.81 & \hspace{2mm}0.201 \\
				\hline
				\hspace{2mm}5 & \hspace{2mm}0.357 & \hspace{2mm}0.370 & \hspace{2mm}0.308 \\
				\hline				
				
			\end{tabular}
		\end{table}
		
\subsection{Weighing face patches}
In the third diagnostic procedure of the multi-path decision, each face patch is used to detect fake or real images. In this section, we will examine the effect of each face patch on detecting fake or real images. 
Therefore, we double the effect of some patches on the diagnosis and give them a weight of two in the final decision.
To reach a fair judgment, we compare different accuracy results such as the accuracy of each face patch, the entire face, the facial patches, and the second diagnostic method based on the concatenation of the feature vectors from each face patch for the fourth and fifth datasets. 
For the fourth dataset, Table \ref{table16} shows the effect of considering weights of the facial patches (mouth, nose, left eye, right cheek, left cheek) and the entire face image  on the accuracy of distinguishing fake images from real ones. 
The weight of the entire face in Table~\ref{table16} is 1 because in the third diagnostic procedure, the full face image is not taken into consideration, in the first diagnostic procedure only does it exist. 
The accuracy of diagnosis on the fourth data set, when the weights of all facial patches are equal to 1, is 77.89\% for the left eye, 76.31\% for the right eye, 85.26\% for the left cheek, 82.63\% for the right cheek, 70.52\% for the chin, 92.10\% for the mouth and 84.21\% for the nose.
According to the accuracy of the facial patches that we examined, and which are presented in Table \ref{table16}, for the fourth data-set, the mouth has the highest diagnostic accuracy with 92.10\%, followed by the left cheek with a diagnostic accuracy of 85.26\%, the nose with a detection accuracy of 84.21\%, the right cheek with a detection accuracy of 82.63\%, and the left eye with a detection accuracy of 77.89\%. Therefore, we try to double the effect of the patches on the face, including the mouth, right cheek, nose, left cheek and left eye. If we double the effect of the left cheek and mouth in the third diagnostic procedure, we obtain a maximum accuracy of 94.21\%, which is about 3.71\% better than when we do not consider the double weight for them. When we double the weight of the right cheek and mouth, we obtain an accuracy of 90.53\%, which is the same as when we didn't consider the weight for the facial patches. The accuracy improves almost 1.6\% when we double the weight for the mouth and nose or the left eye and mouth. 
		\begin{table}[htp]
			
			\begin{center}
				
				\caption{Impact of weight for the fourth dataset}
				\label{table16}
				\renewcommand{\arraystretch}{1}
				\setlength{\extrarowheight}{10pt}
				\scalebox{0.7}{%
					\begin{tabular}{|c|c|c|c|c|c|c|c|c|c|c|c|}
						\toprule
						& Entire Face & Left eye & Right eye & Left cheek & Right cheek & Chin & Mouth & nose & Total & Concatenate & Final\\ \midrule
						weight & 2 & 1 & 1 & 1 & 1 & 1 & 2 & 1 & 1 & 1 & 1 \\
						\hline
						Accuracy & 79.5 & 77.89 & 76.31 & 85.26 & 82.63 & 70.52 & 92.10 & 84.21 & 93.68 & 90.52 & 92.63 \\
						\hline
						weight & 1 & 1 & 1 & 2 & 1 & 1 & 2 & 1 & 1 & 1 & 1 \\
						\hline
						Accuracy & 79.5 & 77.89 & 76.31 & 85.26 & 82.63 & 70.52 & 92.10 & 84.21 & 94.21 & 92.10 & 94.21 \\
						\hline
						weight & 1 & 1 & 1 & 1 & 2 & 1 & 2 & 1 & 1 & 1 & 1 \\
						\hline
						Accuracy & 79.5 & 77.89 & 76.31 & 85.26 & 82.63 & 70.52 & 92.10 & 84.21 & 90.52 & 92.10 & 90.53 \\
						\hline
						weight & 1 & 1 & 1 & 1 & 1 & 1 & 2 & 2 & 1 & 1 & 1 \\
						\hline
						Accuracy & 79.5 & 77.89 & 76.31 & 85.26 & 82.63 & 70.52 & 92.10 & 84.21 & 92.10 & 92.10 & 92.10 \\
						\hline
						weight & 1 & 2 & 1 & 1 & 1 & 1 & 2 & 1 & 1 & 1 & 1 \\
						\hline
						Accuracy & 79.5 & 77.89 & 76.31 & 85.26 & 82.63 & 70.52 & 92.10 & 84.21 & 92.10 & 92.63 & 92.10 \\								
						\hline																	
				\end{tabular}}
			\end{center}
		\end{table}
  
		\begin{table}[htp]
			\begin{center}
				\caption{impact of weight for the fifth dataset}
				\label{table15}
				\renewcommand{\arraystretch}{1.2}
				\setlength{\extrarowheight}{10pt}
				\scalebox{0.7}{%
					\begin{tabular}{|c|c|c|c|c|c|c|c|c|c|c|c|}
						\toprule
						& Entire Face & Left eye & Right eye & Left cheek & Right cheek & Chin & Mouth & nose & Total & Concatenate & Final\\ \midrule
						weight & 2 & 1 & 1 & 1 & 1 & 1 & 2 & 1 & 1 & 1 & 1 \\
						\hline
						Accuracy & 84.04 & 72.07 & 74.64 & 73.78 & 79.77 & 67.52 & 93.15 & 75.49 & 85.47 & 81.48 & 86.32 \\
						\hline
						weight & 1 & 1 & 1 & 1 & 1 & 1 & 2 & 2 & 1 & 1 & 1 \\
						\hline
						Accuracy & 84.04 & 72.07 & 74.64 & 73.78 & 79.77 & 67.52 & 93.15 & 75.49 & 84.04 & 81.48 & 86.89 \\
						\hline
						weight & 1 & 1 & 1 & 1 & 2 & 1 & 2 & 1 & 1 & 1 & 1 \\
						\hline
						Accuracy & 84.04 & 72.07 & 74.64 & 73.78 & 79.77 & 67.52 & 93.15 & 75.49 & 85.18 & 82 & 87.46 \\
						\hline
						weight & 1 & 2 & 1 & 1 & 1 & 1 & 2 & 1 & 1 & 1 & 1 \\
						\hline
						Accuracy & 84.04 & 72.07 & 74.64 & 73.78 & 79.77 & 67.52 & 93.15 & 75.49 & 82.90 & 79.48 & 85.47 \\
						\hline				
						
				\end{tabular}}
			\end{center}
		\end{table}

For the fifth dataset, Table \ref{table15} shows the effects of the weight of each patch and the entire face. The nose, left eye, right cheek and mouth patches are weighed double. The accuracy of diagnosis, when the weights of all facial patches are equal to 1, is 72.07\% for the left eye, 74.64\% for the right eye, 73.78\% for the left cheek, 79.77\% for the right cheek, 67.52\% for the chin, 93.15\% for the mouth and 75.49\% for the nose. As shown in Table \ref{table15}, for the fourth data set, the highest diagnostic accuracy is achieved by the mouth with 93.15 percent, followed by the right cheek with 79.77 percent, and the nose with 75.49\%. Therefore, we try to double the weight of the face patches, including the mouth, right cheek and nose. Additionally, we test the weight of the left eye, which we examined in the fourth dataset, for the fifth dataset. According to the analysis and results of Table \ref{table15} and Table \ref{table16}, the accuracy of mouth detection is higher than other facial patches and the accuracy of chin detection is lower than other facial patches.
In third diagnostic procedure, if we double the weight of the right cheek and mouth, we obtain 87.46\% accuracy, which is almost 2.56\% better than when we do not consider weights for the patches. We achieve an accuracy of 85.47\% when we double the weights for the left eye and mouth, which is almost 0.57\% better than before. We also achieve an accuracy of 86.69\% when we double the weights for the mouth and nose, which is almost 1.79\% better than before.
Finally, we summarize weighing and not weighing facial patches in the third diagnostic approach:
		\begin{itemize}
			\item On average, it increases the final accuracy by 1.84\% and 1.63\% in the fourth and fifth datasets, respectively.
			\item The highest increase of the final accuracy in the fourth and fifth data sets is equal to 3.71\% and 2.5\%, respectively.
			\item The lowest increase of the accuracy in the fourth and fifth data sets is equal to 0.03\% and 0.57\%, respectively.
		\end{itemize}

		\subsection{Modified proposed approach with less time complexity}
		
		\begin{figure}[ht]
			
			\includegraphics[width=\linewidth]{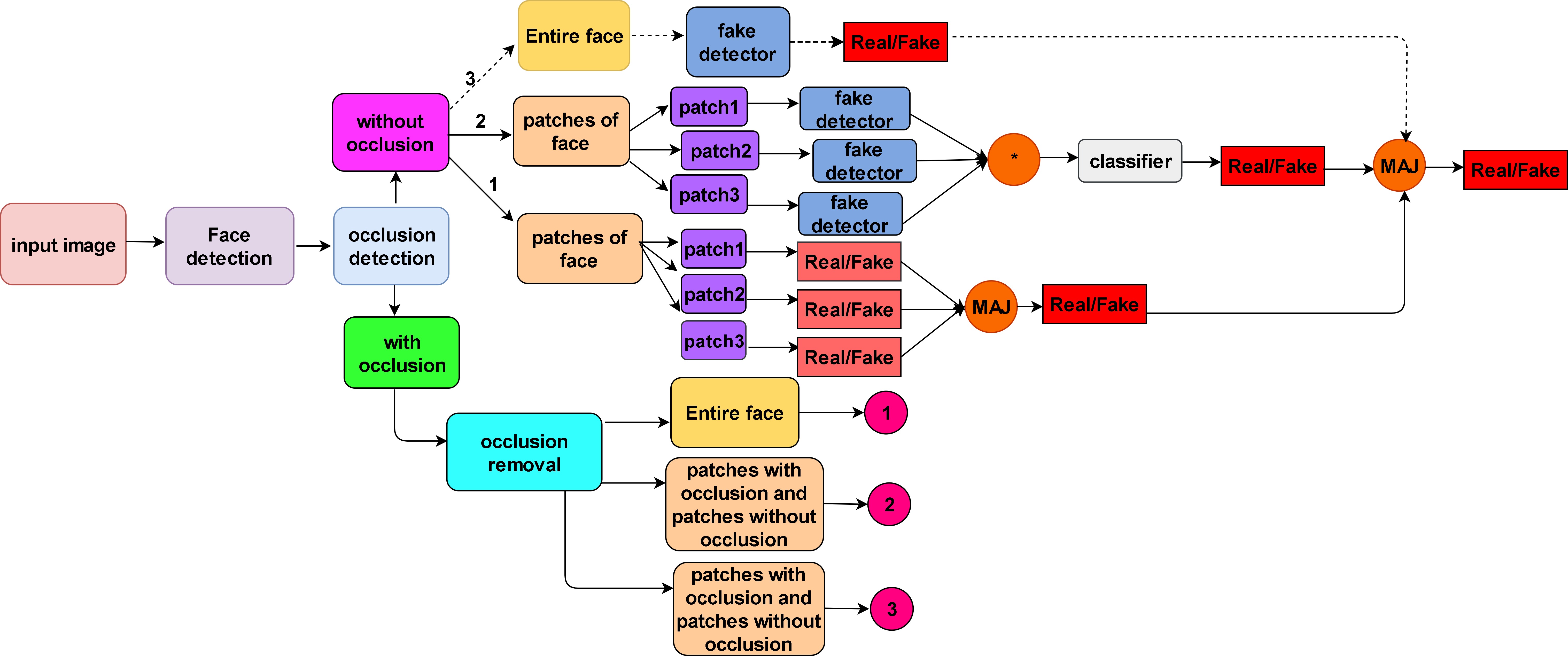}
			\caption{proposed diagram with less time complexity}
			\label{fig_proposed}
		\end{figure}
		
The time complexity of the proposed approach can be reduced, if we first follow the second and third diagnostic procedures, which rely on face patches. Predicted labels can be considered as the final labels if the second diagnostic or the third diagnostic procedures achieves 100\% accuracy. 
Fig.~\ref{fig_proposed} shows that the first diagnostic procedure, shown as a dotted line, does not need to compute if second and third diagnostic procedures, which are patch-based, achieve 100 \% accuracy. This prioritization in computing the procedures also eliminates the maximum voting at the end of the architecture. If we had stopped at the first path to reduce the time complexity, we would have only reached 100\% accuracy for the first dataset. 
		
\section{Conclusion}
We introduced a new patch-based approach to distinguish real images from fake images produced by GAN and examined the occlusion challenge among the deepfake challenges in this study. Moreover, a human study was conducted to compare the performance of human detection with the detection of deep learning models. In our proposed approach, a multi-path decision including three diagnostic procedures was used. One procedure decided based on the entire face and others decided based on face patches. In addition, two types of semantic and block patching were tested to determine the type of final patching of our proposed approach. Additionally, we examined the phase of occlusion removal in the proposed approach. Moreover, we examined the effect of patches on the final diagnosis of whether the image is fake or real by weighing patches. Finally, we introduced a modified version of our approach to reduce temporal complexity.


\newpage
\begin{thebibliography}{99}
	\bibitem{c1}  Z. Liu, X. Qi, and P. H. Torr, "Global texture enhancement for fake face detection in the wild," in Proceedings of the IEEE/CVF Conference on Computer Vision and Pattern Recognition, 2020, pp. 8060-8069.
	\bibitem{c2}  A. Jain, R. Singh, and M. Vatsa, "On detecting gans and retouching based synthetic alterations," in 2018 IEEE 9th International Conference on Biometrics Theory, Applications and Systems (BTAS), 2018: IEEE, pp. 1-7. 
	\bibitem{c3}  A. Jain, P. Majumdar, R. Singh, and M. Vatsa, "Detecting GANs and retouching based digital alterations via DAD-HCNN," in Proceedings of the IEEE/CVF Conference on Computer Vision and Pattern Recognition Workshops, 2020, pp. 672-673.
			
	\bibitem{c4} X. Wu, K. Xu, and P. Hall, "A survey of image synthesis and editing with generative adversarial networks," Tsinghua Science and Technology, vol. 22, no. 6, pp. 660-674, 2017.
	\bibitem{c5} H. Li, B. Li, S. Tan, and J. Huang, "Detection of deep network generated images using disparities in color components. arXiv 2018," arXiv preprint arXiv:1808.07276.
	\bibitem{c6} G. B. Huang, M. Mattar, T. Berg, and E. Learned-Miller, "Labeled faces in the wild: A database forstudying face recognition in unconstrained environments," in Workshop on faces in'Real-Life'Images: detection, alignment, and recognition, 2008. 
	\bibitem{c7} S. McCloskey and M. Albright, "Detecting gan-generated imagery using color cues," arXiv preprint arXiv:1812.08247, 2018.
	\bibitem{c8} T. Karras, T. Aila, S. Laine, and J. Lehtinen, "Progressive growing of gans for improved quality, stability, and variation," arXiv preprint arXiv:1710.10196, 2017.
			
			
	\bibitem{c9}  T.-C. Wang, M.-Y. Liu, J.-Y. Zhu, A. Tao, J. Kautz, and B. Catanzaro, "High-resolution image synthesis and semantic manipulation with conditional gans," in Proceedings of the IEEE conference on computer vision and pattern recognition, 2018, pp. 8798-8807.
	\bibitem{c10} C. Chen, S. McCloskey, and J. Yu, "Focus manipulation detection via photometric histogram analysis," in Proceedings of the IEEE Conference on Computer Vision and Pattern Recognition, 2018, pp. 1674-1682.
	\bibitem{c11} N. Yu, L. S. Davis, and M. Fritz, "Attributing fake images to gans: Learning and analyzing gan fingerprints," in Proceedings of the IEEE/CVF International Conference on Computer Vision, 2019, pp. 7556-7566.
	\bibitem{c12} I. Gulrajani, F. Ahmed, M. Arjovsky, V. Dumoulin, and A. Courville, "Improved training of wasserstein gans," arXiv preprint arXiv:1704.00028, 2017.
	\bibitem{c13} R. Wang et al., "Fakespotter: A simple yet robust baseline for spotting ai-synthesized fake faces," arXiv preprint arXiv:1909.06122, 2019.
	\bibitem{c14} H. Dang, F. Liu, J. Stehouwer, X. Liu, and A. K. Jain, "On the detection of digital face manipulation," in Proceedings of the IEEE/CVF Conference on Computer Vision and Pattern recognition, 2020, pp. 5781-5790. 
	\bibitem{c15} Y.-X. Zhuang and C.-C. Hsu, "Detecting generated image based on a coupled network with two-step pairwise learning," in 2019 IEEE International Conference on Image Processing (ICIP), 2019: IEEE, pp. 3212-3216.
	\bibitem{c16} C.-C. Hsu, C.-Y. Lee, and Y.-X. Zhuang, "Learning to detect fake face images in the wild," in 2018 International Symposium on Computer, Consumer and Control (IS3C), 2018: IEEE, pp. 388-391. 
	\bibitem{c17} L. Nataraj et al., "Detecting GAN generated fake images using co-occurrence matrices," Electronic Imaging, vol. 2019, no. 5, pp. 532-1-532-7, 2019.
	\bibitem{c18} C.-C. Hsu, Y.-X. Zhuang, and C.-Y. Lee, "Deep fake image detection based on pairwise learning," Applied Sciences, vol. 10, no. 1, p. 370, 2020.
	\bibitem{c19} H. Farid, "Image forgery detection," IEEE Signal processing magazine, vol. 26, no. 2, pp. 16-25, 2009.
	\bibitem{c20}  H. Mo, B. Chen, and W. Luo, "Fake faces identification via convolutional neural network," in Proceedings of the 6th ACM Workshop on Information Hiding and Multimedia Security, 2018, pp. 43-47. 
	\bibitem{c21} F. Marra, D. Gragnaniello, D. Cozzolino, and L. Verdoliva, "Detection of gan-generated fake images over social networks," in 2018 IEEE Conference on Multimedia Information Processing and Retrieval (MIPR), 2018: IEEE, pp. 384-389. 
	\bibitem{c22} J. C. Neves, R. Tolosana, R. Vera-Rodriguez, V. Lopes, H. Proença, and J. Fierrez, "Ganprintr: Improved fakes and evaluation of the state of the art in face manipulation detection," IEEE Journal of Selected Topics in Signal Processing, vol. 14, no. 5, pp. 1038-1048, 2020.
	\bibitem{c23} A. Bharati, R. Singh, M. Vatsa, and K. W. Bowyer, "Detecting facial retouching using supervised deep learning," IEEE Transactions on Information Forensics and Security, vol. 11, no. 9, pp. 1903-1913, 2016.
	\bibitem{c24} R. Tolosana, S. Romero-Tapiador, J. Fierrez, and R. Vera-Rodriguez, "Deepfakes evolution: Analysis of facial regions and fake detection performance," in International Conference on Pattern Recognition, 2021: Springer, pp. 442-456. 
	\bibitem{c25} E. Kee and H. Farid, "A perceptual metric for photo retouching," Proceedings of the National Academy of Sciences, vol. 108, no. 50, pp. 19907-19912, 2011.
			
	\bibitem{c26} Z. Akhtar, D. Dasgupta, and B. Banerjee, “Face Authenticity: An Overview of Face Manipulation Generation, Detection and Recognition,” SSRN Electronic Journal, 2019, doi: 10.2139/ssrn.3419272.
			
	\bibitem{c27} R. Tolosana, R. Vera-Rodriguez, J. Fierrez, A. Morales, and J. Ortega-Garcia, "Deepfakes and beyond: A survey of face manipulation and fake detection," arXiv preprint arXiv:2001.00179, 2020.
			
	\bibitem{c28} K. B. Meena and V. Tyagi, "Image forgery detection: survey and future directions," in Data, Engineering and applications: Springer, 2019, pp. 163-194.
	\bibitem{c29} Tero Karras, Samuli Laine, and Timo Aila. “A style based generator architecture for generative adversarial networks”. In: Proceedings of the IEEE/CVF conference on computer vision and pattern recognition. 2019,pp. 4401–4410.
	\bibitem{c30} Ziwei Liu et al. “Deep learning face attributes in the wild”. In: Proceedings of the IEEE international conference on computer vision. 2015, pp. 3730–3738.
	\bibitem{c31} Tero Karras, Samuli Laine, and Timo Aila. “A style based generator architecture for generative adversarial networks”. In: Proceedings of the IEEE/CVF conference on computer vision and pattern recognition. 2019,pp. 4401–4410.
	\bibitem{c32} Yunjey Choi et al. “StarGAN: Unified generative adversarial networks for multi-domain image-to-image translation”. In: Proceedings of the IEEE conference on computer vision and pattern recognition. 2018,pp. 8789–8797.
	\bibitem{c33} Tero Karras, Samuli Laine, Miika Aittala, Janne Hellsten,Jaakko Lehtinen, and Timo Aila. Analyzing and improving the image quality of StyleGAN. In Proceedings of the IEEE Conference on Computer Vision and Pattern Recognition (CVPR), 2020. 1, 2, 4
	\bibitem{c34} Tero Karras, Timo Aila, Samuli Laine, and Jaakko Lehtinen.Progressive growing of gans for improved quality, stability,and variation. arXiv preprint arXiv:1710.10196, 2017.
	\bibitem{c35} Leon Gatys, Alexander S Ecker, and Matthias Bethge. Texture synthesis using convolutional neural networks. In Advances in neural information processing systems, pages 262–270, 2015.
    \bibitem{c36} U. A. Ciftci, I. Demir, and L. Yin, "Fakecatcher: Detection of synthetic portrait videos using biological signals," IEEE transactions on pattern analysis and machine intelligence, 2020. 
		
\end{thebibliography}
\end{document}